\documentclass[manuscript]{acmart}
\AtBeginDocument{%
  \providecommand\BibTeX{{%
    \normalfont B\kern-0.5em{\scshape i\kern-0.25em b}\kern-0.8em\TeX}}}

\setcopyright{acmcopyright}
\copyrightyear{2020}
\acmYear{2020}
\acmDOI{10.1145/1122445.1122456}

\acmPrice{15.00}
\acmISBN{978-1-4503-XXXX-X/18/06}
\usepackage{subfig}
\usepackage{color}
\usepackage[ruled,vlined,linesnumbered]{algorithm2e}
\begin{document}

\title{Iterative Boosting Deep Neural Networks for Predicting Click-Through Rate}


\author{Amit Livne}
\email{livneam@post.bgu.ac.il}
\author{Roy Dor}
\email{rdo@post.bgu.ac.il}
\author{Eyal Mazuz}
\email{mazuze@post.bgu.ac.il}
\author{Tamar Didi}
\email{tamarin@post.bgu.ac.il}
\author{Bracha Shapira}
\email{bshapira@bgu.ac.il}
\author{Lior Rokach}
\email{liorrk@bgu.ac.il}

\affiliation{%
  \institution{Ben-Gurion University of the Negev}
  \city{Beer-Sheva}
  \state{Israel}
}

\renewcommand{\shortauthors}{Livne, et al.}

\begin{abstract}
The click-through rate (CTR) reflects the ratio of clicks on a specific item to its total number of views. It has significant impact on websites' advertising revenue. Learning sophisticated models to understand and predict user behavior is essential for maximizing the CTR in recommendation systems. Recent works have suggested new methods that replace the expensive and time-consuming feature engineering process with a variety of deep learning (DL) classifiers capable of capturing complicated patterns from raw data; these methods have shown significant improvement on the CTR prediction task. While DL techniques can learn intricate user behavior patterns, it relies on a vast amount of data and does not perform as well when there is a limited amount of data. We propose XDBoost, a new DL method for capturing complex patterns that requires just a limited amount of raw data. XDBoost is an iterative three-stage neural network model influenced by the traditional machine learning boosting mechanism. XDBoost's components operate sequentially similar to boosting; However, unlike conventional boosting, XDBoost does not sum the predictions generated by its components. Instead, it utilizes these predictions as new artificial features and enhances CTR prediction by retraining the model using these features. Comprehensive experiments conducted to illustrate the effectiveness of XDBoost on two datasets demonstrated its ability to outperform existing state-of-the-art (SOTA) models for CTR prediction.
\end{abstract}



\newcommand{\roy}[1]{{\textcolor{blue}{[Roy: #1]}}}
\newcommand{\amit}[1]{{\textcolor{red}{[Amit: #1]}}}
\newcommand{\bracha}[1]{{\textcolor{orange}{[Bracha: #1]}}}
\newcommand{\lior}[1]{{\textcolor{green}{[Lior: #1]}}}
\newcommand{\Robin}[1]{{\textcolor{purple}{[Robin: #1]}}}

\maketitle

\section{Introduction}
\label{sec:introduction}
Over the past decade, the popularity and use of the Internet have rapidly increased. Each day, the number of users and mobile devices utilizing this technology grows. Surfing the Internet is one of the most commonly performed activities, with users visiting websites via their devices countless times throughout the day.
Click-through rate (CTR) is a critical aspect of every web page, and many recommender engines within such sites aim at recommending to visitors the next item in order to maximize CTR. CTR is defined as the ratio of clicks on a specific link to its total number of impressions. Many recommender systems (RSs) suggest a ranked recommendation list to a user. The RS needs to sort the list of recommendations to maximize the chance for positive interaction within the user (i.e., click). RS sort the items within the recommendation list by their estimated CTR. Additionally, in other applications such as online advertising, improving CTR is important for increasing the revenue. Thus, each CTR estimation is adjusted by the benefit the system receives for each candidate within the recommendation list. 
Many studies in the recommender systems domain propose various methods for improvement the accuracy of predicting the click-through for e-commerce sites or ads \cite{cheng2016wide,juan2016field,richardson2007predicting}.

The authors of \cite{rendle2010factorization} introduced factorization machines (FMs) in which features are transformed into embedding space, and pairwise feature interactions are modeled as an inner product of those embedded representations. In \cite{blondel2016higher}, the authors suggested extending FMs to capture higher-order feature interactions. Guo et al. \cite{guo2017deepfm} presented, DeepFM, A model that combines the power of FM and deep learning to learn both low and high order feature interactions for solving the CTR prediction task. Features play a crucial role in the success of many predictive systems. However, using raw features rarely leads to optimal results. Thus, data scientists spend a lot of effort generating new features to improve predictive models \cite{he2014practical}.

Due to improvements in computational power, we have recently witnessed the emergence of new methodologies that consider deep learning techniques \cite{guo2017deepfm,huang2019fibinet,lian2018xdeepfm} to solve the CTR prediction task compared to traditional approaches \cite{bauman2010ctr,rendle2010factorization}. 
% \bracha{how doe sth following paragraph relates to the former. You start talkiung about factorization machines... you must do some connection - such as saying that recently factorization machines were suggested as a leading methods for CTR prediction, and then start talking about FM - there must be a flow in a paper...}
Moreover, these studies suggest skipping the feature generation phase and including it as part of a DL mechanism that captures the most suitable features automatically. However, these DL algorithms rely on training with a large amount of data to artificially generate features and may not perform as well when using only a small amount of data.
To address this limitation, we suggest a new iterative boosting deep neural network (DNN) algorithm, XDBoost, that automatically crafting artificial features using a limited amount of data. 
Boosting refers to ensemble mechanism that combine several weak learners into a strong learner \cite{drucker1994boosting}. Its main idea is training predictors sequentially, where each predictor tries to correct its predecessor. In adaptive boosting (AdaBoost), suggested by \cite{freund1995desicion}, a base classifier (i.e., decision tree) is trained first; this classifier is used to make predictions on the training set, and this is followed by increasing the relative weights of misclassified training instances. In \cite{chen2016xgboost}, the authors presented XGBoost, a scalable tree-based machine learning system, in which the method suggested by \cite{friedman2000additive} was modified, improving the regularized objective. CatBoost, presented by \cite{prokhorenkova2018catboost}, is an innovative boosting algorithm for processing categorical features.
Our proposed XDBoost method involves the generation of new features that capture the estimated error distribution using a limited amount of data iteratively. The iterative boosting mechanism aids XDBoost to achieve more accurate CTR prediction.
% Thus, allowing us to improveWithin each iteration of XDBoost's boosting, we predict the estimated error and incorporate it as a new feature for the model. distribution should increase the density of the error distribution near the zero value and result in more accurate CTR prediction .\newline

Our main contributions are summarized as follows:
\begin{enumerate}
    \item We propose a new neural network model, XDBoost, that integrates a boosting mechanism within state of the art (SOTA) DNN to address the CTR prediction task using limited amount of data. Incorporating estimated error via boosting mechanism within DNN allowing XDBoost to improve its CTR predictions. 
    \item We evaluate XDBoost on various datasets, including a public CTR prediction dataset and a proprietary real-world dataset, demonstrating consistent improvement on existing SOTA models for CTR prediction. 
    \item We analyze XDBoost sensitivity to the size of training data and compare the performance of XDBoost to other SOTA models when there is a limited amount of training data available. We show that XDBoost is especially beneficial for scenarios with limited amount of data available.
    % \item \bracha{We evaluate the ability of XDBoost and its base classifier, DeepFM, to provide a dense error distribution around zero. XDBoost outperformed its base classifier and yielded a denser error distribution around zero.bracha - I don't understand why this a contribution and what is the contribution whwy would I care if the error distribution is arund zero}
    \item We explore XDBoost's performance in addressing the cold start problem for new items. XDBoost outperforms all of the baselines.
\end{enumerate}
The rest of this paper is structured as follows: Section \ref{sec:related_work} describes related work, and in Section \ref{sec:method} we present our proposed neural network XDBoost. In Section \ref{sec:experiment} we describe our evaluation and present the results. Finally, in Section \ref{sec:discission} we discuss the results and in Section \ref{sec:conclusion} we provide concluding remarks and discuss future work.

\section{Related Work}
\label{sec:related_work}
%  In this section we review related work on subjects which are closely related to the method we present in this paper.
% In this paper, a new iterative boosting DNN is proposed for CTR prediction. The most related domains are CTR prediction and
% boosting, and in this section, we discuss related work in these two domains.
\subsection{CTR Prediction}
Development of the Internet and mobile devices in recent years has increased the importance of CTR prediction. As a result, many studies have been performed to try to maximize CTR prediction capabilities. CTR prediction is the task of predicting the probability that user $u$ will click on item $i$ in a given context $c$. CTR prediction is essential for businesses that rely on the pay-per-click (PPC) model. The two SOTA methods aimed at this task described below are based on a combination of traditional methods and neural network-based methods. 

The first is factorization machine (FM) \cite{rendle2010factorization}, a traditional method used to capture interactions between features that became very popular after the Netflix Prize competition \cite{bennett2007netflix, bell2007lessons, bell2007bellkor, takacs2008matrix}. FM has been proven successful in many domains including: computer vision \cite{guillamet2002non} and recommendation systems \cite{rendle2010factorization}. However, they often struggle to capture complex patterns and nonlinear interactions, as neural networks often do \cite{bishop2006pattern}. 
The second is a novel artificial neural network (ANN) architecture called Wide \& Deep Learning \cite{cheng2016wide} that was created by Google.
Wide \& Deep is utilizing wide linear models and a DNN, combining the benefits of memorization and generalization to capture more complex patterns and interactions.
The Wide \& Deep architecture has had significant impact on recent studies and contributed to the creation of new SOTA models. 
A variant of FM that extended \cite{rendle2010factorization}, field-aware factorization machines (FFM), proposed by Juan et al. \cite{juan2016field}, addressed the task of CTR prediction. Feature engineering, a difficult manual task requiring time and domain experts, can significantly improve model performance \cite{cheng2016wide}. 
Recent studies in the CTR prediction domain suggest using feature extraction methods to avoid manual feature extraction and implicitly generate new features within their models. Guo et al. \cite{guo2017deepfm} presented DeepFM to capture both low and high order features from raw features. Inspired by DeepFM and Wide \& Deep, Lian et al. \cite{lian2018xdeepfm} replaced the FM layer of DeepFM with a novel cross-network they call the compressed interaction network (CIN), to capture feature interactions. Their method, referred to as xDeepFM, aims to learn certain bounded-degree feature interactions explicitly while learning arbitrary low and high order feature interactions implicitly. Huang et al. \cite{huang2019fibinet} used SENET \cite{hu2018squeeze} to dynamically evaluate feature importance, combined with bilinear feature interactions (FiBiNET), feeding them to a DNN for prediction.

In this research, we suggest a new method that considers implicit raw feature interactions to capture complex patterns. Additionally, we build on another traditional machine learning concept, boosting.

\subsection{Boosting}
Boosting is an abstract mechanism for improving the classification by reducing bias and variance usually applied in ensemble methods. A highly accurate prediction rule is found by combining rough and moderately inaccurate rules of thumb, which are called weak learners. A weak learner is defined as a classifier that is only slightly correlated with the true classification. This theory is based on Valiant's probably approximately correct (PAC) learning model \cite{valiant1984theory}. 
The main variation between many boosting algorithms is their method of weighting training data points and hypotheses, used to create a set of simple rules that have high variance between them.

The most basic boosting algorithm is AdaBoost \cite{freund1995desicion}, which has undergone intense theoretical study and empirical testing. AdaBoost uses a set of decision trees called a forest, to perform prediction. It generates one tree at a time, calculates its prediction error, and gives each training instance a different weight. Then, the next tree is generated differently, due to the weight changes of the instances. Newer boosting methods suggested using gradient boosting factorization machines to incorporate a feature selection algorithm with FM \cite{cheng2014gradient}. For the CTR prediction task, Ling et al. \cite{ling2017model} suggested an ensemble method using DNN and the gradient boosting decision tree (GBDT) algorithm, and several other studies have suggested methods that use neural network-based FM combined with GBDT \cite{zhou2018novel, wang2019novel}.

XGBoost \cite{chen2016xgboost} is currently considered the SOTA boosting algorithm. It has gained popularity in the machine learning community due to its fast performance which stems from its use of parallelization and hardware optimization. XGBoost is based on GBDT, and it uses regularization by penalizing more complex models to prevent overfitting. It handles different types of sparsity patterns by learning the best missing values depending on training loss, and uses the weighted quantile sketch technique to find the optimal split points effectively.
% \bracha{whay is catboost relevant? do you use it?}
% \amit{it is a different boosting method, we consider it as a baseline. It is a good baseline because it handles categorical features differently than XGBOOST}
CatBoost \cite{prokhorenkova2018catboost} is a boosting method that is designed to handle categorical features. CatBoost generates new features from all available categorical feature combinations and utilizes them in the GBDT method for classification.
Although most boosting algorithms were designed for ensemble methods, recent work has tried using the boosting concept on neural networks. Roy et al. \cite{roy2017error} used boosting techniques to fine-tune a CNN image segmentation class weights in cases where labeled data is limited for solving the multi-class classification task. Notably, they change and update the loss function. Mosca and Magoulas \cite{mosca2017deep} embraced the use of boosting techniques in DNNs by using transfer learning \cite{pan2009survey} to quickly generate DNN weak learners. However, they build on multiple classifiers and not a \textbf{single} classifier as we do.

In our model, we suggest an iterative mechanism similar to gradient boosting, however instead of summing the boosted prediction, we reuse the estimated error prediction as a feature for retraining our model. By using the estimated error prediction as a feature, we are able to fuse DNN with boosting.
A detailed description of our model is provided in Section \ref{sec:method}.

\section{Method}
\label{sec:method}
We aim to design an iterative boosting DNN architecture for predicting the CTR. Specifically, we present a network that learns the estimated errors during the training phase and incorporates it iteratively as inputs to the network. Knowing the estimated errors and incorporate them within the model allows us to learn the relation between these error values and true labels and therefore improve the CTR predictions. 
% \bracha{ I would expecft here an intution of wht incorporating the error is a good thing why would it work- you go into technical very fast but dont explain the intuition}\amit{so should i explain it again like i did at the end of the introduction?}.

\subsection{Problem Formulation}
\label{subsec:problem_formulation}
The input for the CTR prediction task 
is a set $D$ composed of $N$ quartettes. Each quartette $(u, i, c, Y)\in D$ denotes an interaction event where user $u$ was exposed to an item $i$ while considering contextual side information $c$ regarding this interaction. $c$ includes additional information about $u$ and $i$ in two different representations: categorical fields (i.e., content category or campaign language) and continuous fields (i.e., quality level). $Y\in\{0,1\}$ is the associated label indicating user click behaviors (Y=1 indicates that user $u$ clicked on item $i$ under contextual side information $c$, Y=0 otherwise). The CTR prediction task is building a
prediction model to estimate the probability for user $u$ clicking a specific item $i$ in a given context $c$.

\subsection{XDBoost}
The proposed method, XDBoost, is based on an iterative three-stage neural network model build on SOTA deep learning classifier (DLC) for CTR Prediction task. XDBoost constructed of a \textbf{single} DLC and \textbf{several} deep learning regressors (DLRs) that operate sequentially. DLC predictions are determined by the sigmoid function. DLR, is identical SOTA component as DLC. However, DLR aims to \textit{estimate the error produced by the DLC's predictions}. Since the CTR is between the range of zero to one, its error distribution can vary from -1 to +1. Thus, instead of the sigmoid activation used in DLC, DLR uses hyperbolic tangent activation. 
% mentioned in equation \ref{eq:tanh}. 
XDBoost' DLRs share the same structure. Each DLR is dedicated to a specific iteration of XDBoost.

\begin{figure}
    \centering
    \includegraphics[scale=0.5]{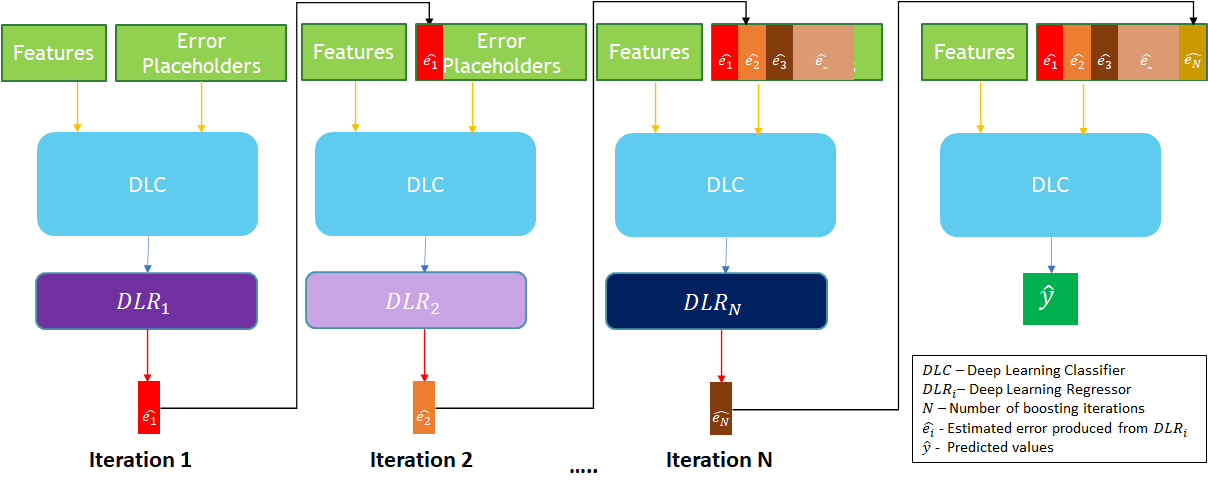}
    \caption{The architecture of our proposed XDBoost model}
    \label{fig:XDBoost}
\end{figure}

XDBoost's goal is to derive a learning model that can learn feature interactions in an end-to-end manner without any feature engineering besides raw features. To use XDBoost, we use raw features that includes the user ID and item ID. We create $N$ additional empty features that we call error placeholders. The estimated error distribution of the DLC in iteration $i$ is derived by $DLR_i$, and it populates the empty placeholder $error_i$. XDBoost includes a single DLC classifier instance and $N$ DLR instances, and can be seen in Figure \ref{fig:XDBoost}. 

 Moreover, XDBoost has two properties: the number of boosting iterations defined by $N$ and the error learning rate multiplier factor defined by $E_{LR}$. A brief description of iteration $i$ consist of three stages is provided below:
% \bracha{a figure describing the architecture that you can refer to it in the text is VERY!!! missing}
\begin{enumerate}
    \item The DLC, an existing SOTA classifier, aims to predict the CTR.
    \item A new DLR instance $DLR_i$, learns the classifier's estimated error distribution $error_i$. 
    \item The DLC uses the error estimation provided by $error_i$ as an input, instead of the corresponding empty placeholder feature.
\end{enumerate}

We describe XDBoost's creation, training, and prediction steps in the algorithms that follow.
In order to create an instance of XDBoost, we apply algorithm \ref{algo:create_XDBoost}.

%%%%%%%%%%%%%%%%  Creating XDBoost %%%%%%%%%%%%%%%%%%%%%%%%%
\begin{algorithm}

\SetKw{KwBy}{by}
\KwIn{$N$ : number of boosting iterations (scalar), an integer scalar greater than 0.\\
$E_{LR}$ : error learning rate multiplier factor, a float scalar bounded by a predefined range $[0,1]$\\
$DLC$ : a SOTA DLC for CTR prediction, a tensorflow model.\\
\\  }
\KwOut{$XDBoost$ : a new instance of XDBoost}

$XDBoost_{N} \leftarrow N $ \\
$XDBoost_{E_{LR}} \leftarrow E_{LR} $ \\
$XDBoost_{DLC} \leftarrow DLC\left (  \right )$\\
\For{$i$ in $XDBoost_{N}$}
{
 $XDBoost_{DLR_i} \leftarrow DLR_i\left (  \right )$\\
}
\Return $XDBoost$ 
 \caption{Creating XDBoost}
 \label{algo:create_XDBoost}
\end{algorithm} 
%%%%%%%%%%%%%%%%  END - Creating XDBoost %%%%%%%%%%%%%%%%%%%%%%%%%
We initialize the properties of XDBoost in lines 1-3. In line 4, we create a new instance of DLC. Lines 4-5 describe the creation of XDBoost's DLRs. Last, we return a new untrained instance of $XDBoost$ in line 6. XDBoost's training consists of two steps and performed sequentially as described in Algorithm \ref{algo:train_XDBoost}. First, in lines 1-3, we create artificial zero-based features; the number of features is based on the number of boosting iterations declared while initializing XDBoost, defined by $XDBoost_N$. Second, in lines 4-11 we train XDBoost by applying the iterative process. Each iteration $i$ is based on the following three stages. First stage described in lines 5-6: train $XDBoost_{DLC}$ and generate its predictions. Second stage described in lines 7-9: estimating the $XDBoost_{DLC}$ error distribution utilizing $XDBoost_{DLR_i}$. Third stage described in line 10-11: populate artificial feature $i$ and retrain $XDBoost_{DLC}$. Last, in line 12 we return the trained instance of $XDBoost$.

% In line 5 and 6 we train  the Each iteration $i$ includes the following components defined by $XDBoost_{DLC}$ and $XDBoost_{DLR_i}$. The training of these components . \newline

%%%%%%%%%%%%%%%%  Training XDBoost %%%%%%%%%%%%%%%%%%%%%%%%%
\begin{algorithm}

\SetKw{KwBy}{by}
\KwIn{$X_{train}$: train set features; $\overrightarrow{y_{train}}$ : train set binary target label data\\ $XDBoost$ : an instance of XDBoost
\\  }

% \KwOut{$\overrightarrow{\hat{y_{test}}}$ : predicted target label for test-set}
\KwOut{$XDBoost$: trained boosted instance of XDBoost}
% \emph{Preliminary step: creating artificial zero-based features }\\

\For{$i$ in $XDBoost_N$}
{
$\overrightarrow{\hat{error_i}} \leftarrow 0 $\\
 $X_{train}[-XDBoost_{N}+i] \leftarrow \overrightarrow{\hat{error_i}}$  \\
}
% \emph{Training}\\
\For{$i$ in $XDBoost_N$}
{
    % \textbf{\textit{First stage: train DLC classifier}}\\
    $XDBoost_{DLC}.fit(X_{train} , {\overrightarrow{y_{train}}})$\\
    $\overrightarrow{\hat{y}_{DLC}} \leftarrow XDBoost_{DLC}.predict(X_{train})$\\
    $\overrightarrow{error_{DLC}} \leftarrow ({\overrightarrow{y_{train}}} - \overrightarrow{\hat{y}_{DLC}})$
    
    %  \textbf{\textit{Second stage: estimating error utilizing $DLR_i$}}\\
     $XDBoost_{DLR_i}.fit(X_{train} , \overrightarrow{error_{DLC}})$\\
     $\overrightarrow{\hat{error_{train}}} \leftarrow XDBoost_{DLR_i}.predict(X_{train})$\\
    %  $\overrightarrow{\hat{error_{test}}} \leftarrow DLR_i.predict(X_{test})$\\
     
    %   \textbf{\textit{Third stage: populate artificial feature and retrain DLC}}\\
     $X_{train}[-XDBoost_{N}+i]\leftarrow(XDBoost_{E_{LR}}*\overrightarrow{\hat{error_{train}}})$\\
    %  $X_{test}[-I+i]\leftarrow(E_{LR}*\overrightarrow{\hat{error_{test}}}*E_{LR})$\\
     $XDBoost_{DLC}.fit(X_{train},\overrightarrow{{y_{train}}})$\\
}
    % $\overrightarrow{\hat{y_{test}}} = DLC.predict(X_{test})$\\
% \Return $\overrightarrow{\hat{y_{test}}}$\\
\Return $XDBoost$
 \caption{Training XDBoost}
 \label{algo:train_XDBoost}
\end{algorithm}
%%%%%%%%%%%%%%%%  END - Traning XDBoost %%%%%%%%%%%%%%%%%%%%%%%%%

In contrast to traditional classifiers, generating predictions via XDBoost includes two steps that operate iteratively. In each iteration, the estimated error is predicted in the first step, and the second step populates the artificial features for the test set. Algorithm \ref{algo:predict_XDBoost} provides a full description of how XDBoost generates predictions. First step described in line 2: estimating error utilizing $XDBoost_{DLR_i}$.
Second step described in line 3: populate artificial feature $i$ in the test set.
%%%%%%%%%%%%%%%%  Predicting XDBoost %%%%%%%%%%%%%%%%%%%%%%%%%

\begin{algorithm}[h]
\KwIn{$XDBoost$ : XDBoost model; $X_{test}$: test set features;\\  }

% \KwOut{$\overrightarrow{\hat{y_{test}}}$ : predicted target label for test-set}
\KwOut{$\overrightarrow{\hat{y_{test}}}$ : $XDBoost$ CTR prediction}

% $N \leftarrow XDBoost_N$     \\
\For{$i$ in $XDBoost_N$}
{
    % \textbf{\textit{First stage - DLC predictionr}}\\
    % $\overrightarrow{\hat{y}_{DLC}} \leftarrow XDBoost_{DLC}.predict(X_{test})$\\
    % $\overrightarrow{error_{DLC}} \leftarrow ({\overrightarrow{y_{test}}} - \overrightarrow{\hat{y}_{DLC}})$
    
    %  \textbf{\textit{First stage - estimating error utilizing $XDBoost_{DLR_i}$}}\\
    %  $DLR_i.fit(X_{train} , \overrightarrow{error_{DLC}})$\\
     $\overrightarrow{\hat{error_{test}}} \leftarrow XDBoost_{DLR_i}.predict(X_{test})$\\
    %  $\overrightarrow{\hat{error_{test}}} \leftarrow DLR_i.predict(X_{test})$\\
     
    %  \textbf{\textit{Second stage - predict DLC}}\\
    
     $X_{test}[-XDBoost_N+i]\leftarrow(XDBoost_{E_{LR}}*\overrightarrow{\hat{error_{test}})}$\\
}
    $\overrightarrow{\hat{y_{test}}} = XDBoost_{DLC}.predict(X_{test})$\\
\Return $\overrightarrow{\hat{y_{test}}}$\\
 \caption{Predicting via XDBoost}
 \label{algo:predict_XDBoost}
\end{algorithm}

Last, in line 4-5, we generate $XDBoost$ predictions utilizing $XDBoost_{DLC}$ and return its predictions.
%%%%%%%%%%%%%%%%  END - Predicting XDBoost %%%%%%%%%%%%%%%%%%%%%%%%%
% \roy{I think this algorithm also need to initial placeholders with 0}

\section{EXPERIMENTS}
\label{sec:experiment}
In this Section, we describe the extensive experiments conducted to answer the
following research questions (RQs):

\begin{itemize}
    % \item (RQ1) How does XDBoost's performance compare to the performance of SOTA DLC, when examining a portion of the available training set without any feature engineering besides raw features? Notably, does XDBoost outperform its base classifier?
    % \bracha{ I would like to suggest another wording- Does XDBoost' boosting mechanism improves the performance of SOTA DLC on raw features. Notabely, does XDBoost outperform its base classifier, specifically we would like to investigate this question for various portion of the data to examine the effect of small training sets  on the performance of XDBOOST vs. SOTA classifiers} 
    \item (RQ1) Does XDBoost's boosting mechanism improve the performance of SOTA DLC on raw features? Notably, does XDBoost outperform its base classifier? Specifically, we would like to investigate this question for various portions of the data to examine the effect of small training sets on the performance of XDBoost vs. SOTA DLC.
    \item (RQ2) Does XDBoost's boosting mechanism improve the performance of SOTA boosting on raw features? Notably, we would like to investigate the effect of XDBoost performance when examining different portions of the available training set.
   
    \item (RQ3) One of the major challenges in recommendation systems is the cold start problem \cite{ricci2015recommender}. In classic recommendation systems, the cold start problem occurs when new users or items which may not have any ratings at all are added to the system. How does XDBoost's ability to generalize and address the cold start problem compare to the abilities of SOTA deep learning algorithms for CTR prediction? How do different portions of the training set influence those results?
    
\end{itemize}

The experiments described later in the paper will be conducted in order to address these questions.

\subsection{Experimental Settings}
\subsubsection{Datasets}
\label{subsec:datasets}
We evaluate the effectiveness of our proposed method on the following datasets:

\begin{itemize}
    
    \item \textbf{Taboola:} Taboola is an advertising company that provides advertisements, such as the "Around the Web" and "Recommended for You" boxes at the bottom of many online news articles. Taboola provides approximately 450 billion article recommendations each month for more than a billion unique users. The Taboola dataset consists of a sample of 15 days of ad click-through data which is ordered chronologically. It contains click logs with 34 million data instances. Each instance consists of 26 fields which reflect the elements of a single ad impression.
    % \amit{add info about the fields}
    \item \textbf{Avauz:} The Avazu dataset is widely used in
many CTR model evaluation. It consists of several days of ad click-through data which is ordered chronologically. It contains click logs with 40 million data instances. For each instance, there are 24 fields which reflect the elements of a single ad impression. The Avazu dataset is publicly accessible.\footnote{Avazu dataset: http://www.kaggle.com/c/avazu-ctr-prediction}
\end{itemize}

Table \ref{tab:dataset_summary} provides a summary of the datasets used in this study.

\begin{table}[h]
\centering
\caption{Dataset Summary}
\begin{tabular}{@{}lll@{}}
\toprule
                        & \multicolumn{1}{c}{Taboola} & \multicolumn{1}{c}{Avazu} \\ \midrule
\ Number of records           & 34M                         & 40M                       \\
\ Number of unique ads        & 94K                         & 40M                       \\
\ Number of unique users      & 16M                         & Unknown                    \\
CTR (\%)                  & 49.9                        & 15.2                      \\
Period of sample (days) & 15                          & 11                        \\
Number of fields & 26                          & 24                        \\\bottomrule
\end{tabular}
\label{tab:dataset_summary}
\end{table}
\subsubsection{Evaluation Metrics}
We use two evaluation metrics in our experiments: the area under the ROC curve (\textbf{AUC}) and the \textbf{Log Loss}. Both metrics are widely used to evaluate CTR prediction performance \cite{guo2017deepfm,huang2019fibinet,liu2019feature}. 
\begin{itemize}
    \item \textbf{AUC:} Area under ROC curve is a widely used metric in evaluating classification problems. The upper bound of the AUC is one, and the larger the AUC, the better. 
    \item \textbf{Log Loss:} Takes into account the uncertainty of your prediction based on how much it varies from the actual label. The lower bound of the log loss is zero, indicating that the two distributions match perfectly, and a smaller value indicates better performance.
\end{itemize}
\subsubsection{Baselines}
\label{subsec:baselines}

In order to test the proposed method, and to answer the research questions we conducted a series of offline simulations and compared them to SOTA deep learning algorithms that are designed specifically to address the CTR prediction task. Additionally, because we integrate boosting mechanism in our method, we compare it to other SOTA algorithms that incorporate boosting mechanism.
Specifically, we use the following algorithms as baselines:
\begin{enumerate}  

\item \textit{SOTA deep learning algorithms:}
    \begin{itemize}
        \item \textbf{\textit{DeepFM}} \cite{guo2017deepfm}, a DNN model that integrates the architectures of FM and DNNs. It models low order feature interactions like FM and models high order feature interactions like DNNs. 
    % \item \textbf{\textit{xDeepFM}} \cite{lian2018xdeepfm}, a neural network model that extends the DeepFM architecture. It replaces the FM layer of DeepFM with a novel cross-network referred to as a compressed interaction network (CIN), to capture feature interactions.
    \item \textbf{\textit{xDeepFM}} \cite{lian2018xdeepfm}, a DNN model suggesting to learn certain bounded-degree feature interactions explicitly combined with low and high order feature interactions implicitly.
    \item \textbf{\textit{FiBiNET}} \cite{huang2019fibinet}, a DNN model suggesting a new way to calculate the feature interactions using bilinear function.
    % \item \textbf{\textit{FGCNN}} \cite{liu2019feature}, a DNN that consists of two components: feature generation utilizing a CNN and a classifier.
    \end{itemize}
\item \textit{SOTA boosting algorithms:}
    \begin{itemize}
        \item \textbf{\textit{XGBoost}} \cite{chen2016xgboost}, an open source software library \footnote{XGBoost: https://xgboost.readthedocs.io/en/latest/} which provides a gradient boosting model. 
    \item \textbf{\textit{CatBoost}} \cite{prokhorenkova2018catboost}, an open source software library \footnote{CatBoost: https://catboost.ai/} which provides a high performance gradient boosting model.
    \end{itemize}
\end{enumerate}
For the implementation of DeepFM, xDeepFM, and FiBiNET we use \textit{deepctr} open source package.\footnote{deepctr: https://github.com/shenweichen/DeepCTR}
\subsubsection{Train-Validation-Test Split}
\label{subsec:train-validation-test-split}
In \cite{huang2019fibinet,liu2019feature}, the authors demonstrated their proposed methods on several datasets, using random splits. However, since %including the Avazu dataset. 
% \bracha { this is not nice to say... you have to say that although in somestudies they authors decided to split the data .... we prefered to split it ... so that it better simulates the real wordl scenario and prevents data leakage}
both of the datasets used in our study \ref{subsec:datasets} (Taboola and Avazu) are sorted chronologically. We split the data using the timestamp. This split simulates the real-world scenario, as real systems train on available data up to a certain timestamp and predict for the following days. Such split therefore prevents data leakage. We split each dataset as follows: the most recent 20\% of the interactions are considered the test set. Of the remaining 80\% of the data, the latest 8\% is used as a validation set and 72\% is used for training.
\subsubsection{Sub-Training Sets}
\label{subsec:sub_train_splits}
\begin{figure}[h]
    \centering
    \includegraphics[width=0.8\textwidth]{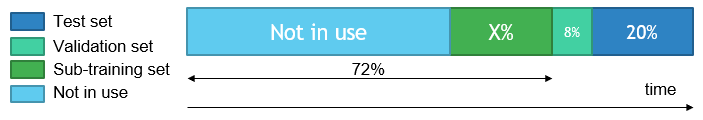}
    \caption{Splitting data into subsets}
    \label{fig:Sub-Traning}
\end{figure}

To address all RQs that relate to the influence of the size of the dataset, i.e. training the model using only a portion of the available training set, we created "sub" training sets from all of the available data while preserving chronologically constant validation and test sets, 8\% and 20\% of the data, respectively (as described in \ref{subsec:train-validation-test-split}). We separate all available training data into two components: the last X\% is considered a sub-training set, and the remaining data is not in use. An illustration of splitting the data into sub-training sets is provided in Figure \ref{fig:Sub-Traning}. In the figure it can be observed that the range of all available training data (72\%) consists of two parts: the sub-training set (X\%) and the portion that is not in use. X can vary from 1\% (indicating that a very small number of instances is used in the training process) to 72\% (indicating that all of the available training data is used).

We used several different sub-training sets. Specifically, we examine the performance of the following sub-training sets sizes: 1\%, 5\%, 10\%, 20\%, 40\%, 60\% and 72\% while preserving the validation and test set constants.
\subsubsection{Class Distributions}
\label{sec:class_dist}
We conduct our experiments on two datasets, Taboola and Avazu (as described in Section \ref{subsec:datasets}). While the Taboola dataset label data is balanced, the Avazu dataset label data is unbalanced; specifically, its CTR is 15.2\%, which creates a situation in which non-click instances are more dominant than click instances. To deal with this imbalance ratio and treat both classes equally, we set a weight for each class; in this way, we are able to mimic a balanced distribution. Thus, we increase the weights of the click class. The weights are determined by the distribution of the training set as follows: 

\begin{equation}
Class Weights= 
\begin{cases}
$non-clicks$ & 1.0\\
$clicks$ & \frac{Count(training-set_{non-clicks})}{Count(training-set_{clicks})} > 1.0
\end{cases}
\end{equation}

% \begin{equation}
% \begin{split}
%     W_{Y==0} = 1 ;  W_{Y==1} = \frac{Count(training-set_{Y==0})}{Count(training-set_{Y==1})} > 1
% \end{split}
% \end{equation}
\subsubsection{Hyper Parameter tuning}
In our experiments, we implement XDBoost with TensorFlow.\footnote{Tensorflow: https://www.tensorflow.org} The dimension of the embedding layer is set to 64. For the optimization we use the Adam optimizer \cite{kingma2014adam} with a mini-batch size of 1,024, and the learning rate is set to 0.0001. For all deep models, the layer depth is set to three, and all activation functions are ReLU, except for the last activation which is sigmoid. The last activation in each DLR component is tanh. The number of neurons per layer is 128 for the Avazu dataset and 256 for the Taboola dataset. For classification components (i.e., DLC and all baselines), we use binary cross-entropy as the loss. For DLR components, we use the mean absolute error (MAE) as the loss, since MAE is not sensitive to outliers. We conduct our experiments using several RTX 2080 TI GPUs.

% \subsection{Results}
% \label{sec:results}

\subsection{Results}
In order to address our research questions, we generate several variants of XDBoost. Each variant is based on a different SOTA deep learning classifier for CTR prediction. Specifically, we suggest the following XDBoost variants: \textbf{XDBoost-DeepFM}, \textbf{XDBoost-xDeepFM}, and \textbf{XDBoost-FiBiNET}.

\subsubsection{XDBoost Performance Compared to SOTA DLC Using Different Sub-Training Sets (RQ1)} 
\begin{figure}[h]
    \centering
    \includegraphics[scale=0.45]{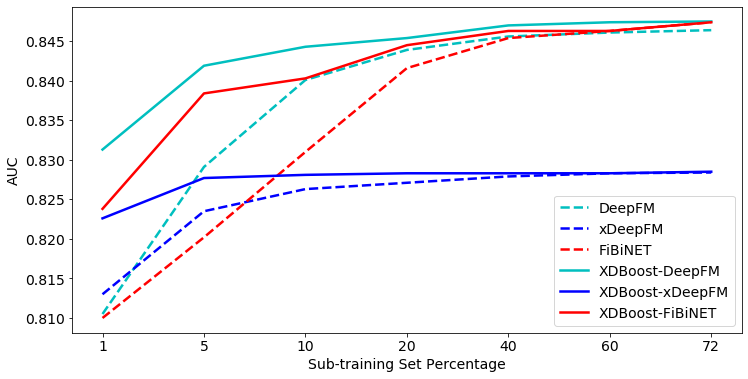}
    \caption{XDBoost' variants AUC performance compared to its base classifier --- Taboola test set.}
    \label{fig:RQ2_AUC_Taboola}
\end{figure}

\begin{figure}[h]
    \centering
    \includegraphics[scale=0.45]{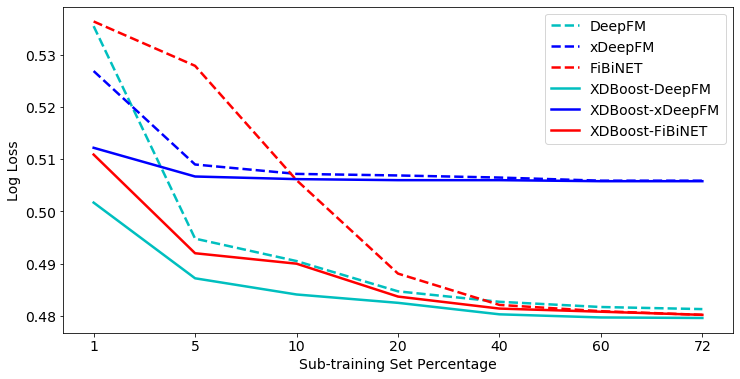}
    \caption{XDBoost' Variants log loss performance compared to its base classifier --- Taboola test set.}
    \label{fig:RQ2_LogLoss_Taboola}
\end{figure}

We compare the performance of our boosted algorithms to their non-boosted variants. In addition, the effectiveness of SOTA DLCs is reduced when using a small amount of data, and SOTA boosting algorithms do not need a lot of data to achieve high scores. Therefore, we explore the effectiveness of the suggested variants of XDBoost also compared to SOTA DLC baselines using different sub-training sets of various dataset sizes. Figure \ref{fig:RQ2_AUC_Taboola} and \ref{fig:RQ2_LogLoss_Taboola} present the AUC and log loss performance on Taboola test set while training on sub-training sets; 72\% indicating that all of the available training data is used. As observed, all DLC algorithms using our XDBoost method outperform their base non-boosted classifiers.

\begin{figure}[h]
    \centering
    \includegraphics[scale=0.45]{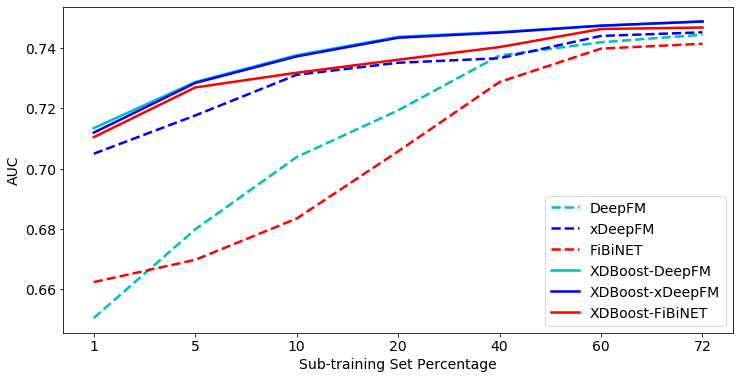}
    \caption{XDBoost' Variants AUC performance compared to its base classifier --- Avazu test set.}
    \label{fig:RQ2_AUC_Avazu}
\end{figure}

\begin{figure}[h]
    \centering
    \includegraphics[scale=0.45]{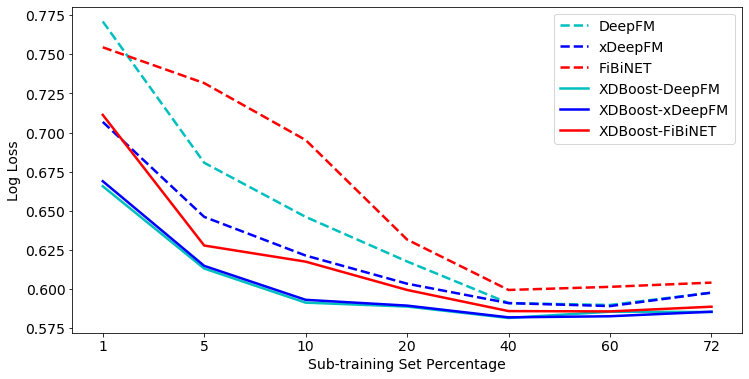}
    \caption{XDBoost' Variants log loss AUC performance compared to its base classifier --- Taboola test set.}
    \label{fig:RQ2_LogLoss_Avazu}
\end{figure}

% When using the Taboola test set, for the AUC metric each XDBoost variant outperforms its SOTA classifier. - we can drop this sentence...
For the Taboola dataset, XDBoost-DeepFM outperforms all baselines in terms of AUC. The improvement ranges between 2.25\% to 2.62\% for 1\% sub-training set compared to other SOTA DLCs. Among the variants of XDBoost, XDBoost-DeepFM results consistently with the best results. When comparing the log loss for different sub-training sets, we can observe trends similar to those of the AUC. For both metrics, XDBoost variants outperform their base classifiers, when using small sub-training sets. However, its impact decreases when using sub-training sets greater than 40\% (i.e., improvements vary between 0.15 and 0.13\%). Notably, when using 60\% and 72\% sub-training sets, more than 20M records are used for training. Thus, allowing DLCs to fulfill their potential.
Figures \ref{fig:RQ2_AUC_Avazu} and \ref{fig:RQ2_LogLoss_Avazu} present the AUC and log loss performance on Avazu test set while training on sub-training sets.

When using the Avazu test set, XDBoost-DeepFM outperforms all baselines, in terms of both the AUC and log loss. XDBoost-DeepFM achieves much better results than its base classifier, DeepFM. For instance, obtained an AUC rate of 0.7378, XDBoost-DeepFM needed only 10\% of the available data. However, to achieve similar results (0.7375), DeepFM needed to use 40\% of the available data. For both metrics, XDBoost variants outperform their base classifiers, when using small sub-training sets. However, similar to the results with the Taboola dataset, their impact decreases when using sub-training sets greater than 40\%.
In order to determine if our proposed boosing achieves significant improvement over the baselines, we conducted the following statistical tests. We first used the adjusted Friedman test in order to reject the null hypothesis that all classifiers perform the same and then the apply the Bonferroni–Dunn test to examine whether our solution algorithm performs significantly better than existing baselines.
The null-hypothesis %that all classifiers perform the same was rejected using the Friedman test
with a confidence level of 99\% for both datasets for all sub-training sets. We proceeded with the Bonferroni-Dunn test and found that \emph{all} the variations of our suggested XDBoost significantly outperform all baselines with 99\% confidence level for both datasets when considering sub-training sets smaller than 60\%. For larger sub-training sets (i.e., 60\% and 72\%), XDBoost-DeepFM and XDBoost-FiBiNET statistically outperform all baselines with a 99\% confidence level for both dataset as well.
%\bracha{what not - xdeepfm? why????}
%\amit{yes, XDBoost variantion of xdeepfm is not sagnificant. why? i assume its because i added it last week and the param tune is far from optimal. however i cannot write it, was it higher anyway}

\subsubsection{XDBoost Performance Compared to SOTA Boosting Algorithms Using Different Sub-Training Sets (RQ2)}
SOTA boosting algorithms do not need a lot of data to achieve high scores. We explore the effectiveness of the suggested variants of XDBoost compared to SOTA boosting algorithms using different sub-training sets. The results of Taboola and Avazu datasets are presented in Tables \ref{tab:RQ2-Taboola} and \ref{tab:RQ2-Avazu} respectively. The best results in each column are denoted in bold. We use the Friedman and Bonferroni-Dunn tests as described in RQ1. Results that are statistically significant (p < 0.01) are denoted by an asterisk (*). As observed, all DLC-boosted algorithms outperform significantly the none- DLC boosting algorithms,and the XDBoost-DeepFM performed best. 

\begin{table*}[h]
\centering
\caption{Comparison of SOTA Boosting Algorithms to XDBoost Variants - Taboola Dataset}
\resizebox{\textwidth}{!}{%
\begin{tabular}{lccccccc|ccccccc}
 & \multicolumn{7}{c|}{AUC} & \multicolumn{7}{c}{Log Loss} \\ \cline{2-15} 
\multicolumn{1}{l|}{\begin{tabular}[c]{@{}l@{}}Models\textbackslash\\ Sub-Training\end{tabular}} & 1\% & 5\% & 10\% & 20\% & 40\% & 60\% & 72\% & 1\% & 5\% & 10\% & 20\% & 40\% & 60\% & 72\% \\ \hline
% \multicolumn{1}{l|}{XGBoost} & 78.02& 76.47 & 75.57 & 75.22 &74.01 & 74.78 & 74.50 & .561	&.573	&.579	&.581&	.590&	.585&	.588 \\
% \multicolumn{1}{l|}{CatBoost} & 79.36&	78.63& 78.15& 77.66& 76.98 & 77.40 & 77.40 & .544&	.549&	.550&	.557&	.563&	.558&	.561 \\ \hline
\multicolumn{1}{l|}{XGBoost} & .815&	.815&	.816&	.813&	.814&	.814	& .815 & .524	&.524&	.523&.526&	.525&	.525&	.523
 \\
\multicolumn{1}{l|}{CatBoost}& .830&	.831&	.832&	.832&	.832&	.831&	.832& .502&.501&.500&.500&.500&.501&.499

  \\ \hline
\multicolumn{1}{l|}{XDBoost-DeepFM} & \textbf{.831*}&	\textbf{.842*}&	\textbf{.844*}&	\textbf{.845*}&	\textbf{.847*}&	\textbf{.847*}&	\textbf{.848*}&
\textbf{.500*}	&\textbf{.487*}&\textbf{.484*}&\textbf{.482*}&\textbf{.480*}&\textbf{.479*}&\textbf{.479*}

  \\ 
\multicolumn{1}{l|}{XDBoost-xDeepFM} & .823*&	.828*&	.828*&	.828*&	.828*&	.828*&	.829*& .512*&.507*&.506*&.506*&.506*&.506*&.506*

 \\ 
\multicolumn{1}{l|}{XDBoost-FiBiNET} & .823*&	.838*&	.840*&	.844*&	.846*&	.846*&	.847*& .511*&.492*&.490*&.484*&.481*&.481*&.480*

 \\ 
\end{tabular}
}

\label{tab:RQ2-Taboola}
\end{table*}

\begin{table*}[h]
\centering
\caption{Comparison of SOTA Boosting Algorithms to XDBoost Variants - Avazu Dataset}
\resizebox{\textwidth}{!}{%
\begin{tabular}{lccccccc|ccccccc}
 & \multicolumn{7}{c|}{AUC} & \multicolumn{7}{c}{Log Loss} \\ \cline{2-15} 
\multicolumn{1}{l|}{\begin{tabular}[c]{@{}l@{}}Models\textbackslash\\ Sub-Training\end{tabular}} & 1\% & 5\% & 10\% & 20\% & 40\% & 60\% & 72\% & 1\% & 5\% & 10\% & 20\% & 40\% & 60\% & 72\% \\ \hline
% \multicolumn{1}{l|}{XGBoost} & 78.02& 76.47 & 75.57 & 75.22 &74.01 & 74.78 & 74.50 & .561	&.573	&.579	&.581&	.590&	.585&	.588 \\
% \multicolumn{1}{l|}{CatBoost} & 79.36&	78.63& 78.15& 77.66& 76.98 & 77.40 & 77.40 & .544&	.549&	.550&	.557&	.563&	.558&	.561 \\ \hline
\multicolumn{1}{l|}{XGBoost} & .676&	.698&	.696&	.696&	.696&	.699&	.701&
.727&	.684&.681&	.679&	.647&	.634&	.629

 \\
\multicolumn{1}{l|}{CatBoost}& .678	& .706	& .712	&.727	&.728	&.728&	.726&
.725	&.671	&.693	&.653	&.622	&.627	&.620

  \\ \hline
\multicolumn{1}{l|}{XDBoost-DeepFM} & \textbf{.714*}	&\textbf{.729*}&	\textbf{.738*}&	\textbf{.744*}&	\textbf{.746*}&	\textbf{.748*}&	\textbf{.749*}& \textbf{.666*}&	\textbf{.613*}&	\textbf{.591*}&	\textbf{.588*}&	\textbf{.582*}&	\textbf{.583*}&	\textbf{.585*}
  \\ 
\multicolumn{1}{l|}{XDBoost-xDeepFM} & .712*&	.728*&	.737*&	.743*&	.745*&	.747*&	.749&
.669*&	.615*&.593*&	.589*&	.582*&	.584*&	.586
 \\ 
\multicolumn{1}{l|}{XDBoost-FiBiNET} & .711*&	.727*&	.732*&	.736*&	.741*&	.747*&	.747*&
.711*&	.628*&	.618*	&.599*&.586*&.586*&.589*
 \\ 
\end{tabular}
}
\label{tab:RQ2-Avazu}
\end{table*}

When using the Taboola dataset, XDBoost-DeepFM outperforms all baselines for all sub-training sets. As we assumed, traditional tree-based boosting algorithms (i.e., XGBoost and CatBoost) perform well when using a small amount of data. However, they cannot model complex nonlinear interactions like DNNs.

When using the Avazu dataset, XDBoost-DeepFM is much more effective compared to XGBoost and CatBoost. XDBoost-DeepFM outperforms all baselines by at least 5.08\% in terms of the AUC when training on 1\% sub-training set. 

\subsubsection{Cold Start (RQ3)}
\begin{figure}[h]
    \centering
    \includegraphics[scale=0.6]{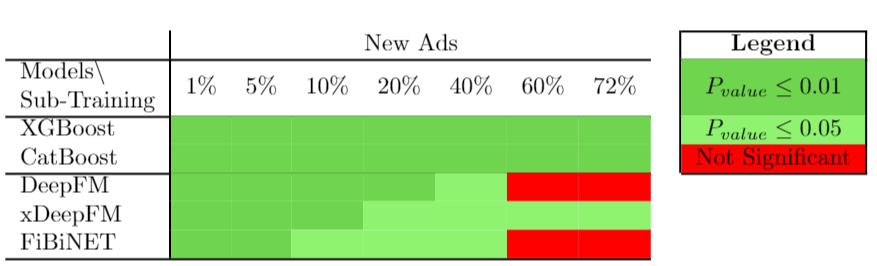}
    \caption{Significant level conducted by Bonferroni–Dunn test of XDBoost-DeepFM compared to each baseline for cold-start items.}
    \label{fig:sagnificant}
\end{figure}

% One of the major challenges in recommendation systems is the cold start problem \cite{ricci2015recommender}. In classic recommendation systems, the cold start problem occurs when new users or items which may not have any ratings at all are added to the system.\bracha{up to here should go to the section where you describe the RQ}
To examine the cold start handling of our method we used only the Taboola Dataset. In the Avazu dataset, each instance represents a unique ad ID (item) without any information regarding the user ID. Thus, we can conclude that this dataset is, by definition, a user- cold start dataset, thus, we could not test the effect of the portion of the cold start users on results. However, this is not the case for the Taboola dataset \ref{tab:dataset_summary}. 
% Although the Taboola dataset provides information regarding the user ID, most of the users consist of only a single instance. 
% Therefore, in this RQ we aim to explore the item cold start problem in the Taboola dataset for the SOTA deep learning models suggested.
In this RQ we focus on cold start problem for new ads. Notably, we investigate the effect of XDBoost performance when examining different portions of sub-training sets.
A cold start ad is defined as an ad that was not appearing the the training set. Thus, we filter from the test set items which appear in the training set, resulting in a smaller test set with fewer records within the originally test set. We use several different sub-training sets, and therefore each sub-training require a corresponding smaller test set.
For example, when training our model on the 1\% sub-training set, we filter the ads that appear in the 1\% sub-training set from the original test set (i.e., 6M records) resulting in a smaller test set (i.e., 186K). Then we evaluated every model on the smaller test set. When using greater sub-training sets, the amount of records within the corresponding filtered test set decrease dramatically. 
We conducted several experiments on the Taboola dataset to explore the cold start problem with different sub-training sets. We compared all baselines to our suggested XDBoost-DeepFM, which yields the best results in RQ1 and RQ2. \autoref{tab:cold-ad-id} present the results regarding new ads. The best results in each column are denoted in bold. Results that are statistically significant compared to \emph{all} baseline with significance level of 95\% and 99\% are denoted by an asterisk (*) and two asterisks (**) respectively. We use the Friedman and Bonferroni-Dunn tests as described previously. Moreover, we mark the significant level of XDBoost-DeepFM compared to all baselines with color in \autoref{fig:sagnificant}.

\begin{table*}[h]
\centering
\caption{Comparison of all baselines algorithms to XDBoost-DeepFM addressing the cold start when Examining New Ads.}
\resizebox{\textwidth}{!}{%
\begin{tabular}{lccccccc|ccccccc}
 & \multicolumn{7}{c|}{AUC} & \multicolumn{7}{c}{Log Loss} \\ \cline{2-15} 
\multicolumn{1}{l|}{\begin{tabular}[c]{@{}l@{}}Models\textbackslash\\ Sub-Training\end{tabular}} & 1\% & 5\% & 10\% & 20\% & 40\% & 60\% & 72\% & 1\% & 5\% & 10\% & 20\% & 40\% & 60\% & 72\% \\ \hline
\multicolumn{1}{l|}{XGBoost} & .780& .764 & .755 & .752 & .740 & .747 & .745 & .561	&.573	&.579	&.581&	.590&	.585&	.588 \\
\multicolumn{1}{l|}{CatBoost} & .793 &	.786 & .781& .776& .769 & .774 & .774 & .544&	.549&	.550&	.557&	.563&	.558&	.561 \\ \hline
\multicolumn{1}{l|}{DeepFM} & .771&	.763& .776	& .785	& .793	& .799& .797 & .577&	.576&	.561&	.549&	.538&	.534	& .534 \\
\multicolumn{1}{l|}{xDeepFM} & .758	& .765&	.763&	.762& .754	&.759&.760 & .582	&.574	&.575	&.576	&.580	&.577	&.578 \\
\multicolumn{1}{l|}{FiBiNET}& .782	&.791	&.791	& .797&	.795	&.799	&.799 &.560&	.537&	.541&	.539	&.537&	.531	&.531 \\ \hline
% \multicolumn{1}{l|}{FGCNN}& 70.71&	72.05&	75.16& 76.50	&76.14	&77.01	&76.58 & .963&	.736&	.591&	.568&	.569&.563&	.567 \\ \hline
\multicolumn{1}{l|}{XDBoost-DeepFM} & \textbf{.798**}	&\textbf{.804**}&	\textbf{.799*}&	\textbf{.800*}	&\textbf{.797*}&\textbf{.800}	&\textbf{.801} &\textbf{.541**}&	\textbf{.532**}&	\textbf{.536*}&	\textbf{.534*}&	\textbf{.535}&	\textbf{.530}&	\textbf{.530}\\

% \multicolumn{1}{l|}{{XDBoost-xDeepFM}} &.779& .776& .768& .764& .756& .760 & .761& .562&.564& .572& .574 & .581&.578&.578\\

% \multicolumn{1}{l|}{XDBoost-FiBiNET} &.788& .798& .794&.798&\textbf{.799}& .799 & .800 & .551 &.540 &.542& .535&\textbf{.533}&\textbf{.530}&\textbf{.530}

\end{tabular}
}
\label{tab:cold-ad-id}
\end{table*}

We can observe from~\autoref{tab:cold-ad-id} that XDBoost-DeepFM outperforms all other SOTA models in all cases, in terms of both the AUC and log loss. While XDBoost excels for all sizes of sub-training sets, it has less impact sub-training set of size 40\% and greater. As we can observe from~\autoref{fig:sagnificant} comparing to SOTA boosting algorithms, XDBoost-DeepFM significant level remains 99\% for all sub-training sets. In contrast, the significant confidence level of XDBoost-DeepFM decrease as the size of sub-training sets increase comparing to SOTA DLCs. Notably, XDBoost-DeepFM is not significant only when using 60\% \& 72\% sub-training sets (i.e., more than 20M records for training) and only by comparing it to DeepFM and FiBiNET. 

\section{Discussion}
\label{sec:discission}

% Our experiments show that XDBoost outperforms all other SOTA models on two datasets on both the AUC and log loss. For RQ2, we demonstrate XDBoost's robust performance and its ability to perform better than other SOTA models when considering a limited amount of data. The smaller amount of training data used for training the greater XDBoost improve comparing to other SOTA models. This phenomenon occurs since XDBoost learns the error distribution and incorporate it as a feature within the model iteratively.
% The experiment conducted for RQ3 shows that the XDBoost architecture improves the base model by providing a denser error distribution around zero, demonstrating its effectiveness. When comparing results for the cold start problem, XDBoost's performance on small sub-training sets is emphasized even further. DeepFM and FGCNN are heavily impacted by the sub-training set size, while FiBiNET's performance shows steady improvement over a larger sub-training set. 

Our experiments show that the XDBoost method has improved the performance of SOTA DLCs and SOTA boosting models on two datasets for both AUC and log loss (RQ1 and RQ2) for the CTR prediction task. On both datasets, we can see the impact of XDBoost is increasing when using smaller sub-training sets. Thus, it can be useful, especially when the available training data is limited. The effectiveness of SOTA DLCs is reduced when using a small amount of data. In contrast, SOTA boosting algorithms do not need a lot of data to achieve high scores. XDBoost exploits the traits of both. It does not require a lot of data to achieve high scores; it is improving when using a limited amount of data. Thus, its superiority over DLCs is mostly significant in small training data scenarios which is a very common challenge since in many cases training data is expensive or difficult to collect~\cite{weiss2016survey}.
When compared to SOTA tree-based boosting algorithms (RQ2) on balance dataset (i.e., Taboola), we can observe that these algorithms' performance barely improve even while considering small sub-training sets. In contrast, XDBoost keeps improving with greater sub-training sets. On the other hand, while observing results on unbalanced data (i.e., Avazu dataset) improvement can be seen for both SOTA tree-based boosting algorithms for bigger training sets.
We assume the improvement of CatBoost is significant (7\% improvement between 1\% to 72\% sub-training sets) on the Avazu dataset compared to XGBoost (3.3\% improvement) and XDBoosted models (5\% improvement) because the Avazu dataset consists of binary and categorical features only, hence it is more suitable for CatBoost that is superior in handling such data. 
% Although it requires training multiple times and use multiple regressors, when considering small sub-training sets, the training time and complexity is marginal.
XDBoost can significantly increase DNN models performance, especially when training data is limited. Thus, we recommend adopting the XDBoost architecture when using neural networks and the available data for training is limited in order to maximize the model's performance.
We have demonstrated XDBoost on three examples of DNN algorithms. However, it can be applied easily to other DNNs as well. 
When examining the results it is important to note that a small improvement in the offline AUC is likely to lead to a significant increase in online CTR. As reported in \cite{cheng2016wide}, compared with logistic regression, Wide \& Deep improves the offline AUC by 0.275\%, and the improvement of the online CTR is 3.9\%. Thus, even a minor improvements can be greatly beneficial.

\section{Conclusions}

\label{sec:conclusion}
In this paper, we propose XDBoost, an iterative boosted DNN architecture for predicting the CTR relying on raw features only. XDBoost learns the estimated error rate during the training phase and incorporate it iteratively as an input to the network. This allows learning of the relation between these error values and the true label. Extensive experiments conducted on two large-scale datasets show consistent improvement over existing SOTA models for CTR prediction. XDBoost's effectiveness compared to other SOTA methods increases when there is a limited amount of data.
While exploring the cold start problem for new items, XDBoost outperformed all other baselines in terms of both the AUC and log loss.
In the future, we plan to mix different SOTA regressors in the XDBoost architecture. Additionally, we aim to generalize XDBoost in order to solve other recommendation tasks, such as ranking and rating prediction.

\bibliographystyle{ACM-Reference-Format}
\bibliography{main}


\begin{thebibliography}{34}

%%% the Web address.

\ifx \showCODEN    \undefined \def \showCODEN     #1{\unskip}     \fi
\ifx \showDOI      \undefined \def \showDOI       #1{#1}\fi
\ifx \showISBNx    \undefined \def \showISBNx     #1{\unskip}     \fi
\ifx \showISBNxiii \undefined \def \showISBNxiii  #1{\unskip}     \fi
\ifx \showISSN     \undefined \def \showISSN      #1{\unskip}     \fi
\ifx \showLCCN     \undefined \def \showLCCN      #1{\unskip}     \fi
\ifx \shownote     \undefined \def \shownote      #1{#1}          \fi
\ifx \showarticletitle \undefined \def \showarticletitle #1{#1}   \fi
\ifx \showURL      \undefined \def \showURL       {\relax}        \fi
% invisible to TeX
\providecommand\bibfield[2]{#2}
\providecommand\bibinfo[2]{#2}
\providecommand\natexlab[1]{#1}
\providecommand\showeprint[2][]{arXiv:#2}

\bibitem[\protect\citeauthoryear{Bauman, Kornetova, Topinskiy, and
  Leshiner}{Bauman et~al\mbox{.}}{2010}]%
        {bauman2010ctr}
\bibfield{author}{\bibinfo{person}{K Bauman}, \bibinfo{person}{A Kornetova},
  \bibinfo{person}{V Topinskiy}, {and} \bibinfo{person}{D Leshiner}.}
  \bibinfo{year}{2010}\natexlab{}.
\newblock \showarticletitle{CTR prediction based on click statistic}.
\newblock \bibinfo{journal}{\emph{Machine Learning in Online Advertising}},
  \bibinfo{pages}{8--13}.
\newblock


\bibitem[\protect\citeauthoryear{Bell and Koren}{Bell and Koren}{2007}]%
        {bell2007lessons}
\bibfield{author}{\bibinfo{person}{Robert~M Bell} {and} \bibinfo{person}{Yehuda
  Koren}.} \bibinfo{year}{2007}\natexlab{}.
\newblock \showarticletitle{Lessons from the Netflix prize challenge}.
\newblock \bibinfo{journal}{\emph{Acm Sigkdd Explorations Newsletter}}
  \bibinfo{volume}{9}, \bibinfo{number}{2} (\bibinfo{year}{2007}),
  \bibinfo{pages}{75--79}.
\newblock


\bibitem[\protect\citeauthoryear{Bennett, Elkan, Liu, Smyth, and Tikk}{Bennett
  et~al\mbox{.}}{2007}]%
        {bennett2007netflix}
\bibfield{author}{\bibinfo{person}{James Bennett}, \bibinfo{person}{Charles
  Elkan}, \bibinfo{person}{Bing Liu}, \bibinfo{person}{Padhraic Smyth}, {and}
  \bibinfo{person}{Domonkos Tikk}.} \bibinfo{year}{2007}\natexlab{}.
\newblock \showarticletitle{KDD Cup and Workshop 2007}.
\newblock \bibinfo{journal}{\emph{SIGKDD Explor. Newsl.}} \bibinfo{volume}{9},
  \bibinfo{number}{2}, \bibinfo{pages}{51–52}.
\newblock
\showISSN{1931-0145}
\urldef\tempurl%
\url{https://doi.org/10.1145/1345448.1345459}
\showDOI{\tempurl}


\bibitem[\protect\citeauthoryear{Bishop}{Bishop}{2006}]%
        {bishop2006pattern}
\bibfield{author}{\bibinfo{person}{Christopher Bishop}.}
  \bibinfo{year}{2006}\natexlab{}.
\newblock \bibinfo{booktitle}{\emph{Pattern Recognition and Machine Learning}}.
\newblock \bibinfo{publisher}{Springer}.
\newblock
\urldef\tempurl%
\url{https://www.microsoft.com/en-us/research/publication/pattern-recognition-machine-learning/}
\showURL{%
\tempurl}


\bibitem[\protect\citeauthoryear{Blondel, Fujino, Ueda, and Ishihata}{Blondel
  et~al\mbox{.}}{2016}]%
        {blondel2016higher}
\bibfield{author}{\bibinfo{person}{Mathieu Blondel}, \bibinfo{person}{Akinori
  Fujino}, \bibinfo{person}{Naonori Ueda}, {and} \bibinfo{person}{Masakazu
  Ishihata}.} \bibinfo{year}{2016}\natexlab{}.
\newblock \showarticletitle{Higher-Order Factorization Machines}. In
  \bibinfo{booktitle}{\emph{Proceedings of the 30th International Conference on
  Neural Information Processing Systems}} (Barcelona, Spain)
  \emph{(\bibinfo{series}{NIPS’16})}. \bibinfo{publisher}{Curran Associates
  Inc.}, \bibinfo{address}{Red Hook, NY, USA}, \bibinfo{pages}{3359–3367}.
\newblock
\showISBNx{9781510838819}


\bibitem[\protect\citeauthoryear{Chen and Guestrin}{Chen and Guestrin}{2016}]%
        {chen2016xgboost}
\bibfield{author}{\bibinfo{person}{Tianqi Chen} {and} \bibinfo{person}{Carlos
  Guestrin}.} \bibinfo{year}{2016}\natexlab{}.
\newblock \showarticletitle{XGBoost: A Scalable Tree Boosting System}. In
  \bibinfo{booktitle}{\emph{Proceedings of the 22nd ACM SIGKDD International
  Conference on Knowledge Discovery and Data Mining}} (San Francisco,
  California, USA) \emph{(\bibinfo{series}{KDD ’16})}.
  \bibinfo{publisher}{Association for Computing Machinery},
  \bibinfo{address}{New York, NY, USA}, \bibinfo{pages}{785–794}.
\newblock
\showISBNx{9781450342322}
\urldef\tempurl%
\url{https://doi.org/10.1145/2939672.2939785}
\showDOI{\tempurl}


\bibitem[\protect\citeauthoryear{Cheng, Xia, Zhang, King, and Lyu}{Cheng
  et~al\mbox{.}}{2014}]%
        {cheng2014gradient}
\bibfield{author}{\bibinfo{person}{Chen Cheng}, \bibinfo{person}{Fen Xia},
  \bibinfo{person}{Tong Zhang}, \bibinfo{person}{Irwin King}, {and}
  \bibinfo{person}{Michael~R. Lyu}.} \bibinfo{year}{2014}\natexlab{}.
\newblock \showarticletitle{Gradient Boosting Factorization Machines}. In
  \bibinfo{booktitle}{\emph{Proceedings of the 8th ACM Conference on
  Recommender Systems}} (Foster City, Silicon Valley, California, USA)
  \emph{(\bibinfo{series}{RecSys ’14})}. \bibinfo{publisher}{Association for
  Computing Machinery}, \bibinfo{address}{New York, NY, USA},
  \bibinfo{pages}{265–272}.
\newblock
\showISBNx{9781450326681}
\urldef\tempurl%
\url{https://doi.org/10.1145/2645710.2645730}
\showDOI{\tempurl}


\bibitem[\protect\citeauthoryear{Cheng, Koc, Harmsen, Shaked, Chandra, Aradhye,
  Anderson, Corrado, Chai, Ispir, Anil, Haque, Hong, Jain, Liu, and Shah}{Cheng
  et~al\mbox{.}}{2016}]%
        {cheng2016wide}
\bibfield{author}{\bibinfo{person}{Heng-Tze Cheng}, \bibinfo{person}{Levent
  Koc}, \bibinfo{person}{Jeremiah Harmsen}, \bibinfo{person}{Tal Shaked},
  \bibinfo{person}{Tushar Chandra}, \bibinfo{person}{Hrishi Aradhye},
  \bibinfo{person}{Glen Anderson}, \bibinfo{person}{Greg Corrado},
  \bibinfo{person}{Wei Chai}, \bibinfo{person}{Mustafa Ispir},
  \bibinfo{person}{Rohan Anil}, \bibinfo{person}{Zakaria Haque},
  \bibinfo{person}{Lichan Hong}, \bibinfo{person}{Vihan Jain},
  \bibinfo{person}{Xiaobing Liu}, {and} \bibinfo{person}{Hemal Shah}.}
  \bibinfo{year}{2016}\natexlab{}.
\newblock \showarticletitle{Wide \& Deep Learning for Recommender Systems}. In
  \bibinfo{booktitle}{\emph{Proceedings of the 1st Workshop on Deep Learning
  for Recommender Systems}} (Boston, MA, USA) \emph{(\bibinfo{series}{DLRS
  2016})}. \bibinfo{publisher}{Association for Computing Machinery},
  \bibinfo{address}{New York, NY, USA}, \bibinfo{pages}{7–10}.
\newblock
\showISBNx{9781450347952}
\urldef\tempurl%
\url{https://doi.org/10.1145/2988450.2988454}
\showDOI{\tempurl}


\bibitem[\protect\citeauthoryear{Drucker, Cortes, Jackel, LeCun, and
  Vapnik}{Drucker et~al\mbox{.}}{1994}]%
        {drucker1994boosting}
\bibfield{author}{\bibinfo{person}{Harris Drucker}, \bibinfo{person}{Corinna
  Cortes}, \bibinfo{person}{Lawrence~D Jackel}, \bibinfo{person}{Yann LeCun},
  {and} \bibinfo{person}{Vladimir Vapnik}.} \bibinfo{year}{1994}\natexlab{}.
\newblock \showarticletitle{Boosting and other ensemble methods}.
\newblock \bibinfo{journal}{\emph{Neural Computation}} \bibinfo{volume}{6},
  \bibinfo{number}{6} (\bibinfo{year}{1994}), \bibinfo{pages}{1289--1301}.
\newblock


\bibitem[\protect\citeauthoryear{Freund and Schapire}{Freund and
  Schapire}{1995}]%
        {freund1995desicion}
\bibfield{author}{\bibinfo{person}{Yoav Freund} {and}
  \bibinfo{person}{Robert~E. Schapire}.} \bibinfo{year}{1995}\natexlab{}.
\newblock \showarticletitle{A desicion-theoretic generalization of on-line
  learning and an application to boosting}. In
  \bibinfo{booktitle}{\emph{Computational Learning Theory}},
  \bibfield{editor}{\bibinfo{person}{Paul Vit{\'a}nyi}} (Ed.).
  \bibinfo{publisher}{Springer Berlin Heidelberg}, \bibinfo{address}{Berlin,
  Heidelberg}, \bibinfo{pages}{23--37}.
\newblock
\showISBNx{978-3-540-49195-8}


\bibitem[\protect\citeauthoryear{Friedman, Hastie, Tibshirani,
  et~al\mbox{.}}{Friedman et~al\mbox{.}}{2000}]%
        {friedman2000additive}
\bibfield{author}{\bibinfo{person}{Jerome Friedman}, \bibinfo{person}{Trevor
  Hastie}, \bibinfo{person}{Robert Tibshirani}, {et~al\mbox{.}}}
  \bibinfo{year}{2000}\natexlab{}.
\newblock \showarticletitle{Additive logistic regression: a statistical view of
  boosting (with discussion and a rejoinder by the authors)}.
\newblock \bibinfo{journal}{\emph{The annals of statistics}}
  \bibinfo{volume}{28}, \bibinfo{number}{2} (\bibinfo{year}{2000}),
  \bibinfo{pages}{337--407}.
\newblock


\bibitem[\protect\citeauthoryear{Guillamet and Vitria}{Guillamet and
  Vitria}{2002}]%
        {guillamet2002non}
\bibfield{author}{\bibinfo{person}{David Guillamet} {and}
  \bibinfo{person}{Jordi Vitria}.} \bibinfo{year}{2002}\natexlab{}.
\newblock \showarticletitle{Non-negative Matrix Factorization for Face
  Recognition}. In \bibinfo{booktitle}{\emph{Topics in Artificial
  Intelligence}}, \bibfield{editor}{\bibinfo{person}{M.~Teresa Escrig},
  \bibinfo{person}{Francisco Toledo}, {and} \bibinfo{person}{Elisabet
  Golobardes}} (Eds.). \bibinfo{publisher}{Springer Berlin Heidelberg},
  \bibinfo{address}{Berlin, Heidelberg}, \bibinfo{pages}{336--344}.
\newblock
\showISBNx{978-3-540-36079-7}


\bibitem[\protect\citeauthoryear{Guo, Tang, Ye, Li, and He}{Guo
  et~al\mbox{.}}{2017}]%
        {guo2017deepfm}
\bibfield{author}{\bibinfo{person}{Huifeng Guo}, \bibinfo{person}{Ruiming
  Tang}, \bibinfo{person}{Yunming Ye}, \bibinfo{person}{Zhenguo Li}, {and}
  \bibinfo{person}{Xiuqiang He}.} \bibinfo{year}{2017}\natexlab{}.
\newblock \showarticletitle{DeepFM: a factorization-machine based neural
  network for CTR prediction}.
\newblock \bibinfo{journal}{\emph{arXiv preprint arXiv:1703.04247}}
  \bibinfo{volume}{24}, \bibinfo{number}{3} (\bibinfo{year}{2017}),
  \bibinfo{pages}{262--290}.
\newblock


\bibitem[\protect\citeauthoryear{He, Pan, Jin, Xu, Liu, Xu, Shi, Atallah,
  Herbrich, Bowers, and Candela}{He et~al\mbox{.}}{2014}]%
        {he2014practical}
\bibfield{author}{\bibinfo{person}{Xinran He}, \bibinfo{person}{Junfeng Pan},
  \bibinfo{person}{Ou Jin}, \bibinfo{person}{Tianbing Xu}, \bibinfo{person}{Bo
  Liu}, \bibinfo{person}{Tao Xu}, \bibinfo{person}{Yanxin Shi},
  \bibinfo{person}{Antoine Atallah}, \bibinfo{person}{Ralf Herbrich},
  \bibinfo{person}{Stuart Bowers}, {and} \bibinfo{person}{Joaquin Qui\~{n}onero
  Candela}.} \bibinfo{year}{2014}\natexlab{}.
\newblock \showarticletitle{Practical Lessons from Predicting Clicks on Ads at
  Facebook}. In \bibinfo{booktitle}{\emph{Proceedings of the Eighth
  International Workshop on Data Mining for Online Advertising}} (New York, NY,
  USA) \emph{(\bibinfo{series}{ADKDD’14})}. \bibinfo{publisher}{Association
  for Computing Machinery}, \bibinfo{address}{New York, NY, USA},
  \bibinfo{pages}{1–9}.
\newblock
\showISBNx{9781450329996}
\urldef\tempurl%
\url{https://doi.org/10.1145/2648584.2648589}
\showDOI{\tempurl}


\bibitem[\protect\citeauthoryear{Hu, Shen, and Sun}{Hu et~al\mbox{.}}{2018}]%
        {hu2018squeeze}
\bibfield{author}{\bibinfo{person}{Jie Hu}, \bibinfo{person}{Li Shen}, {and}
  \bibinfo{person}{Gang Sun}.} \bibinfo{year}{2018}\natexlab{}.
\newblock \showarticletitle{Squeeze-and-Excitation Networks}. In
  \bibinfo{booktitle}{\emph{The IEEE Conference on Computer Vision and Pattern
  Recognition (CVPR)}}.
\newblock


\bibitem[\protect\citeauthoryear{Huang, Zhang, and Zhang}{Huang
  et~al\mbox{.}}{2019}]%
        {huang2019fibinet}
\bibfield{author}{\bibinfo{person}{Tongwen Huang}, \bibinfo{person}{Zhiqi
  Zhang}, {and} \bibinfo{person}{Junlin Zhang}.}
  \bibinfo{year}{2019}\natexlab{}.
\newblock \showarticletitle{FiBiNET: Combining Feature Importance and Bilinear
  Feature Interaction for Click-through Rate Prediction}. In
  \bibinfo{booktitle}{\emph{Proceedings of the 13th ACM Conference on
  Recommender Systems}} (Copenhagen, Denmark) \emph{(\bibinfo{series}{RecSys
  ’19})}. \bibinfo{publisher}{Association for Computing Machinery},
  \bibinfo{address}{New York, NY, USA}, \bibinfo{pages}{169–177}.
\newblock
\showISBNx{9781450362436}
\urldef\tempurl%
\url{https://doi.org/10.1145/3298689.3347043}
\showDOI{\tempurl}


\bibitem[\protect\citeauthoryear{Juan, Zhuang, Chin, and Lin}{Juan
  et~al\mbox{.}}{2016}]%
        {juan2016field}
\bibfield{author}{\bibinfo{person}{Yuchin Juan}, \bibinfo{person}{Yong Zhuang},
  \bibinfo{person}{Wei-Sheng Chin}, {and} \bibinfo{person}{Chih-Jen Lin}.}
  \bibinfo{year}{2016}\natexlab{}.
\newblock \showarticletitle{Field-Aware Factorization Machines for CTR
  Prediction}. In \bibinfo{booktitle}{\emph{Proceedings of the 10th ACM
  Conference on Recommender Systems}} (Boston, Massachusetts, USA)
  \emph{(\bibinfo{series}{RecSys ’16})}. \bibinfo{publisher}{Association for
  Computing Machinery}, \bibinfo{address}{New York, NY, USA},
  \bibinfo{pages}{43–50}.
\newblock
\showISBNx{9781450340359}
\urldef\tempurl%
\url{https://doi.org/10.1145/2959100.2959134}
\showDOI{\tempurl}


\bibitem[\protect\citeauthoryear{Kingma and Ba}{Kingma and Ba}{2014}]%
        {kingma2014adam}
\bibfield{author}{\bibinfo{person}{Diederik Kingma} {and}
  \bibinfo{person}{Jimmy Ba}.} \bibinfo{year}{2014}\natexlab{}.
\newblock \showarticletitle{Adam: A Method for Stochastic Optimization}.
\newblock \bibinfo{journal}{\emph{International Conference on Learning
  Representations}} (\bibinfo{date}{12} \bibinfo{year}{2014}).
\newblock


\bibitem[\protect\citeauthoryear{Koren}{Koren}{2009}]%
        {bell2007bellkor}
\bibfield{author}{\bibinfo{person}{Yehuda Koren}.}
  \bibinfo{year}{2009}\natexlab{}.
\newblock \showarticletitle{The bellkor solution to the netflix grand prize}.
\newblock \bibinfo{journal}{\emph{Netflix prize documentation}}
  \bibinfo{volume}{81}, \bibinfo{number}{2009} (\bibinfo{year}{2009}),
  \bibinfo{pages}{1--10}.
\newblock


\bibitem[\protect\citeauthoryear{Lian, Zhou, Zhang, Chen, Xie, and Sun}{Lian
  et~al\mbox{.}}{2018}]%
        {lian2018xdeepfm}
\bibfield{author}{\bibinfo{person}{Jianxun Lian}, \bibinfo{person}{Xiaohuan
  Zhou}, \bibinfo{person}{Fuzheng Zhang}, \bibinfo{person}{Zhongxia Chen},
  \bibinfo{person}{Xing Xie}, {and} \bibinfo{person}{Guangzhong Sun}.}
  \bibinfo{year}{2018}\natexlab{}.
\newblock \showarticletitle{XDeepFM: Combining Explicit and Implicit Feature
  Interactions for Recommender Systems}. In
  \bibinfo{booktitle}{\emph{Proceedings of the 24th ACM SIGKDD International
  Conference on Knowledge Discovery \& Data Mining}} (London, United Kingdom)
  \emph{(\bibinfo{series}{KDD ’18})}. \bibinfo{publisher}{Association for
  Computing Machinery}, \bibinfo{address}{New York, NY, USA},
  \bibinfo{pages}{1754–1763}.
\newblock
\showISBNx{9781450355520}
\urldef\tempurl%
\url{https://doi.org/10.1145/3219819.3220023}
\showDOI{\tempurl}


\bibitem[\protect\citeauthoryear{Ling, Deng, Gu, Zhou, Li, and Sun}{Ling
  et~al\mbox{.}}{2017}]%
        {ling2017model}
\bibfield{author}{\bibinfo{person}{Xiaoliang Ling}, \bibinfo{person}{Weiwei
  Deng}, \bibinfo{person}{Chen Gu}, \bibinfo{person}{Hucheng Zhou},
  \bibinfo{person}{Cui Li}, {and} \bibinfo{person}{Feng Sun}.}
  \bibinfo{year}{2017}\natexlab{}.
\newblock \showarticletitle{Model Ensemble for Click Prediction in Bing Search
  Ads}. In \bibinfo{booktitle}{\emph{Proceedings of the 26th International
  Conference on World Wide Web Companion}} (Perth, Australia)
  \emph{(\bibinfo{series}{WWW ’17 Companion})}.
  \bibinfo{publisher}{International World Wide Web Conferences Steering
  Committee}, \bibinfo{address}{Republic and Canton of Geneva, CHE},
  \bibinfo{pages}{689–698}.
\newblock
\showISBNx{9781450349147}
\urldef\tempurl%
\url{https://doi.org/10.1145/3041021.3054192}
\showDOI{\tempurl}


\bibitem[\protect\citeauthoryear{Liu, Tang, Chen, Yu, Guo, and Zhang}{Liu
  et~al\mbox{.}}{2019}]%
        {liu2019feature}
\bibfield{author}{\bibinfo{person}{Bin Liu}, \bibinfo{person}{Ruiming Tang},
  \bibinfo{person}{Yingzhi Chen}, \bibinfo{person}{Jinkai Yu},
  \bibinfo{person}{Huifeng Guo}, {and} \bibinfo{person}{Yuzhou Zhang}.}
  \bibinfo{year}{2019}\natexlab{}.
\newblock \showarticletitle{Feature Generation by Convolutional Neural Network
  for Click-Through Rate Prediction}. In \bibinfo{booktitle}{\emph{The World
  Wide Web Conference}} (San Francisco, CA, USA) \emph{(\bibinfo{series}{WWW
  ’19})}. \bibinfo{publisher}{Association for Computing Machinery},
  \bibinfo{address}{New York, NY, USA}, \bibinfo{pages}{1119–1129}.
\newblock
\showISBNx{9781450366748}
\urldef\tempurl%
\url{https://doi.org/10.1145/3308558.3313497}
\showDOI{\tempurl}


\bibitem[\protect\citeauthoryear{Mosca and Magoulas}{Mosca and
  Magoulas}{2017}]%
        {mosca2017deep}
\bibfield{author}{\bibinfo{person}{Alan Mosca} {and} \bibinfo{person}{George~D
  Magoulas}.} \bibinfo{year}{2017}\natexlab{}.
\newblock \bibinfo{title}{Deep Incremental Boosting}.
\newblock
\newblock


\bibitem[\protect\citeauthoryear{Pan and Yang}{Pan and Yang}{2009}]%
        {pan2009survey}
\bibfield{author}{\bibinfo{person}{Sinno~Jialin Pan} {and}
  \bibinfo{person}{Qiang Yang}.} \bibinfo{year}{2009}\natexlab{}.
\newblock \showarticletitle{A survey on transfer learning}.
\newblock \bibinfo{journal}{\emph{IEEE Transactions on knowledge and data
  engineering}} \bibinfo{volume}{22}, \bibinfo{number}{10}
  (\bibinfo{year}{2009}), \bibinfo{pages}{1345--1359}.
\newblock


\bibitem[\protect\citeauthoryear{Prokhorenkova, Gusev, Vorobev, Dorogush, and
  Gulin}{Prokhorenkova et~al\mbox{.}}{2018}]%
        {prokhorenkova2018catboost}
\bibfield{author}{\bibinfo{person}{Liudmila Prokhorenkova},
  \bibinfo{person}{Gleb Gusev}, \bibinfo{person}{Aleksandr Vorobev},
  \bibinfo{person}{Anna~Veronika Dorogush}, {and} \bibinfo{person}{Andrey
  Gulin}.} \bibinfo{year}{2018}\natexlab{}.
\newblock \showarticletitle{CatBoost: Unbiased Boosting with Categorical
  Features}. In \bibinfo{booktitle}{\emph{Proceedings of the 32nd International
  Conference on Neural Information Processing Systems}} (Montr\'{e}al, Canada)
  \emph{(\bibinfo{series}{NIPS’18})}. \bibinfo{publisher}{Curran Associates
  Inc.}, \bibinfo{address}{Red Hook, NY, USA}, \bibinfo{pages}{6639–6649}.
\newblock


\bibitem[\protect\citeauthoryear{Rendle}{Rendle}{2010}]%
        {rendle2010factorization}
\bibfield{author}{\bibinfo{person}{Steffen Rendle}.}
  \bibinfo{year}{2010}\natexlab{}.
\newblock \showarticletitle{Factorization Machines}. In
  \bibinfo{booktitle}{\emph{Proceedings of the 2010 IEEE International
  Conference on Data Mining}} \emph{(\bibinfo{series}{ICDM ’10})}.
  \bibinfo{publisher}{IEEE Computer Society}, \bibinfo{address}{USA},
  \bibinfo{pages}{995–1000}.
\newblock
\showISBNx{9780769542560}
\urldef\tempurl%
\url{https://doi.org/10.1109/ICDM.2010.127}
\showDOI{\tempurl}


\bibitem[\protect\citeauthoryear{Ricci, Rokach, and Shapira}{Ricci
  et~al\mbox{.}}{2015}]%
        {ricci2015recommender}
\bibfield{author}{\bibinfo{person}{Francesco Ricci}, \bibinfo{person}{Lior
  Rokach}, {and} \bibinfo{person}{Bracha Shapira}.}
  \bibinfo{year}{2015}\natexlab{}.
\newblock \showarticletitle{Recommender Systems: Introduction and Challenges}.
\newblock In \bibinfo{booktitle}{\emph{Recommender Systems Handbook}},
  \bibfield{editor}{\bibinfo{person}{Francesco Ricci}, \bibinfo{person}{Lior
  Rokach}, {and} \bibinfo{person}{Bracha Shapira}} (Eds.).
  \bibinfo{publisher}{Springer US}, \bibinfo{address}{Boston, MA},
  \bibinfo{pages}{1--34}.
\newblock
\showISBNx{978-1-4899-7637-6}
\urldef\tempurl%
\url{https://doi.org/10.1007/978-1-4899-7637-6_1}
\showDOI{\tempurl}


\bibitem[\protect\citeauthoryear{Richardson, Dominowska, and Ragno}{Richardson
  et~al\mbox{.}}{2007}]%
        {richardson2007predicting}
\bibfield{author}{\bibinfo{person}{Matthew Richardson}, \bibinfo{person}{Ewa
  Dominowska}, {and} \bibinfo{person}{Robert Ragno}.}
  \bibinfo{year}{2007}\natexlab{}.
\newblock \showarticletitle{Predicting clicks: estimating the click-through
  rate for new ads}. In \bibinfo{booktitle}{\emph{Proceedings of the 16th
  international conference on World Wide Web}}. \bibinfo{pages}{521--530}.
\newblock


\bibitem[\protect\citeauthoryear{Roy, Conjeti, Sheet, Katouzian, Navab, and
  Wachinger}{Roy et~al\mbox{.}}{2017}]%
        {roy2017error}
\bibfield{author}{\bibinfo{person}{Abhijit~Guha Roy}, \bibinfo{person}{Sailesh
  Conjeti}, \bibinfo{person}{Debdoot Sheet}, \bibinfo{person}{Amin Katouzian},
  \bibinfo{person}{Nassir Navab}, {and} \bibinfo{person}{Christian Wachinger}.}
  \bibinfo{year}{2017}\natexlab{}.
\newblock \showarticletitle{Error Corrective Boosting for Learning Fully
  Convolutional Networks with Limited Data}. In
  \bibinfo{booktitle}{\emph{Medical Image Computing and Computer Assisted
  Intervention - MICCAI 2017}}, \bibfield{editor}{\bibinfo{person}{Maxime
  Descoteaux}, \bibinfo{person}{Lena Maier-Hein}, \bibinfo{person}{Alfred
  Franz}, \bibinfo{person}{Pierre Jannin}, \bibinfo{person}{D.~Louis Collins},
  {and} \bibinfo{person}{Simon Duchesne}} (Eds.). \bibinfo{publisher}{Springer
  International Publishing}, \bibinfo{address}{Cham},
  \bibinfo{pages}{231--239}.
\newblock
\showISBNx{978-3-319-66179-7}


\bibitem[\protect\citeauthoryear{Tak\'{a}cs, Pil\'{a}szy, N\'{e}meth, and
  Tikk}{Tak\'{a}cs et~al\mbox{.}}{2008}]%
        {takacs2008matrix}
\bibfield{author}{\bibinfo{person}{G\'{a}bor Tak\'{a}cs},
  \bibinfo{person}{Istv\'{a}n Pil\'{a}szy}, \bibinfo{person}{Botty\'{a}n
  N\'{e}meth}, {and} \bibinfo{person}{Domonkos Tikk}.}
  \bibinfo{year}{2008}\natexlab{}.
\newblock \showarticletitle{Matrix Factorization and Neighbor Based Algorithms
  for the Netflix Prize Problem}. In \bibinfo{booktitle}{\emph{Proceedings of
  the 2008 ACM Conference on Recommender Systems}} (Lausanne, Switzerland)
  \emph{(\bibinfo{series}{RecSys ’08})}. \bibinfo{publisher}{Association for
  Computing Machinery}, \bibinfo{address}{New York, NY, USA},
  \bibinfo{pages}{267–274}.
\newblock
\showISBNx{9781605580937}
\urldef\tempurl%
\url{https://doi.org/10.1145/1454008.1454049}
\showDOI{\tempurl}


\bibitem[\protect\citeauthoryear{Valiant}{Valiant}{1984}]%
        {valiant1984theory}
\bibfield{author}{\bibinfo{person}{Leslie~G Valiant}.}
  \bibinfo{year}{1984}\natexlab{}.
\newblock \showarticletitle{A theory of the learnable}.
\newblock \bibinfo{journal}{\emph{Commun. ACM}} \bibinfo{volume}{27},
  \bibinfo{number}{11} (\bibinfo{year}{1984}), \bibinfo{pages}{1134--1142}.
\newblock


\bibitem[\protect\citeauthoryear{Wang, Hu, Lin, and Sun}{Wang
  et~al\mbox{.}}{2019}]%
        {wang2019novel}
\bibfield{author}{\bibinfo{person}{Xiaochen Wang}, \bibinfo{person}{Gang Hu},
  \bibinfo{person}{Haoyang Lin}, {and} \bibinfo{person}{Jiayu Sun}.}
  \bibinfo{year}{2019}\natexlab{}.
\newblock \showarticletitle{A Novel Ensemble Approach for Click-Through Rate
  Prediction Based on Factorization Machines and Gradient Boosting Decision
  Trees}. In \bibinfo{booktitle}{\emph{Web and Big Data}},
  \bibfield{editor}{\bibinfo{person}{Jie Shao}, \bibinfo{person}{Man~Lung Yiu},
  \bibinfo{person}{Masashi Toyoda}, \bibinfo{person}{Dongxiang Zhang},
  \bibinfo{person}{Wei Wang}, {and} \bibinfo{person}{Bin Cui}} (Eds.).
  \bibinfo{publisher}{Springer International Publishing},
  \bibinfo{address}{Cham}, \bibinfo{pages}{152--162}.
\newblock
\showISBNx{978-3-030-26075-0}


\bibitem[\protect\citeauthoryear{Weiss, Khoshgoftaar, and Wang}{Weiss
  et~al\mbox{.}}{2016}]%
        {weiss2016survey}
\bibfield{author}{\bibinfo{person}{Karl Weiss}, \bibinfo{person}{Taghi~M
  Khoshgoftaar}, {and} \bibinfo{person}{DingDing Wang}.}
  \bibinfo{year}{2016}\natexlab{}.
\newblock \showarticletitle{A survey of transfer learning}.
\newblock \bibinfo{journal}{\emph{Journal of Big data}} \bibinfo{volume}{3},
  \bibinfo{number}{1} (\bibinfo{year}{2016}), \bibinfo{pages}{9}.
\newblock


\bibitem[\protect\citeauthoryear{Zhou, Yin, Zhan, Li, Fan, and Jiang}{Zhou
  et~al\mbox{.}}{2018}]%
        {zhou2018novel}
\bibfield{author}{\bibinfo{person}{Feng Zhou}, \bibinfo{person}{Hua Yin},
  \bibinfo{person}{Lizhang Zhan}, \bibinfo{person}{Huafei Li},
  \bibinfo{person}{Yeliang Fan}, {and} \bibinfo{person}{Liu Jiang}.}
  \bibinfo{year}{2018}\natexlab{}.
\newblock \showarticletitle{A Novel Ensemble Strategy Combining Gradient
  Boosted Decision Trees and Factorization Machine Based Neural Network for
  Clicks Prediction}. In \bibinfo{booktitle}{\emph{2018 International
  Conference on Big Data and Artificial Intelligence (BDAI)}}. IEEE,
  \bibinfo{pages}{29--33}.
\newblock
\urldef\tempurl%
\url{https://doi.org/10.1109/BDAI.2018.8546685}
\showDOI{\tempurl}


\end{thebibliography}
\end{document}